\documentclass[11pt,a4paper]{article}
\usepackage[hyperref]{acl2021}
\usepackage{times}
\usepackage{latexsym}
\usepackage[export]{adjustbox}
\usepackage{enumitem}
\usepackage{booktabs}
\usepackage{makecell}
\usepackage{textcomp} 
\usepackage{subcaption}
\usepackage{array}
\usepackage{multirow}

\newcommand{\red}[1]{\textcolor{red}{#1}}
\newcommand{\blue}[1]{\textcolor{blue}{#1}}

\usepackage{microtype}

\aclfinalcopy 
\def\aclpaperid{703} 

\title{TextSETTR: Few-Shot Text Style \\Extraction and Tunable Targeted Restyling}

\author{Parker Riley\textsuperscript{$a$}\thanks{\hspace{2mm}Work done while at Google Research. Please direct correspondence to \texttt{priley3@cs.rochester.edu}, \texttt{nconstant@google.com} and \texttt{xyguo@google.com}.},~
Noah Constant\textsuperscript{$b$},~
Mandy Guo\textsuperscript{$b$},\\
{\bf Girish Kumar\textsuperscript{$c$}\footnotemark[1],}~
{\bf David Uthus\textsuperscript{$b$},}~
{\bf Zarana Parekh\textsuperscript{$b$}}
  \AND
  {\rm\textsuperscript{$a$}University of Rochester} \And
  {\rm\textsuperscript{$b$}Google Research} \And
  {\rm\textsuperscript{$c$}Stanford University}
}

\date{}

\begin{document}
\maketitle
\begin{abstract}
We present a novel approach to the problem of text style transfer.
Unlike previous approaches requiring style-labeled training data, our method makes use of readily-available unlabeled text by relying on the implicit connection in style between adjacent sentences, and uses labeled data only at inference time.
We adapt T5 \citep{2020t5}, a strong pretrained text-to-text model, to extract a style vector from text and use it to condition the decoder to perform style transfer.
As our label-free training results in a style vector space encoding many facets of style, we recast transfers as ``targeted restyling'' vector operations that adjust specific attributes of the input while preserving others.
We demonstrate that training on \emph{unlabeled} Amazon reviews data results in a model that is competitive on sentiment transfer, even compared to models trained fully on labeled data.
Furthermore, applying our novel method to a diverse corpus of unlabeled web text results in a single model capable of transferring along multiple dimensions of style (dialect, emotiveness, formality, politeness, sentiment) despite \emph{no additional training} and using only a handful of exemplars at inference time.

\end{abstract}

\section{Introduction}
\label{intro}
There has been a recent surge of interest in text style transfer, with the aim of training models able to modify specific attributes of input text (e.g.,~sentiment or formality) while preserving the remaining content. For example, a sentiment transfer model might transform the input ``best book ever!\@'' into ``worst book ever!\@'', while a formality transfer model might change the same input into ``This is the best book I have ever read.'' In these contexts, we define ``style'' as the attributes intended to be changed, while ``content'' consists of the attributes intended to be preserved.\footnote{\citet{krishna-etal-2020-reformulating} use a different definition of style, under which certain transfers such as sentiment would instead be examples of \textit{attribute} transfer.}

Work in this area falls into three categories. \textbf{Supervised} approaches like \citet{jhamtani17} transfer between pre-selected styles, and rely on parallel training data to learn the desired input/output correspondence. This method is limited by the availability of parallel corpora. So-called \textbf{``unsupervised''} approaches like \citet{li18} and \citet{lample19} remove the need for parallel data, but still require that all training examples have style labels, and are limited to transfer between a pre-specified set of styles. 
\textbf{Few-shot} approaches like that of \citet{xu20} remove the need for any training labels, instead using a small number of labeled examples during inference.
While the most challenging, this offers the potential for transferring between arbitrary styles at inference time and has significant value, as curated datasets are not available for many style attributes.

In this work, we explore the hypothesis that large pretrained text-to-text models like T5 \citep{2020t5} already contain a strong representation of textual style, which can be extracted and used to condition the decoder of a style transfer model through a relatively lightweight fine-tuning procedure. To isolate style information in the absence of labels, we rely on the observation that style is a ``slow-moving'' feature, which tends to be consistent over large spans of text. Specifically, given two adjacent sentences from an unlabeled corpus, we train our model to extract a ``style vector'' from the first and use that vector to perform denoising and other reconstruction tasks on the second. This technique extends the approach of \citet{lample19} to the few-shot setting, and is loosely reminiscent of the work of \citet{akama-etal-2018-unsupervised}, who found large context windows useful for encoding style information in word embeddings.
Our approach also allows us to reformulate the style transfer operation as a directional operation in style vector space using the difference between target and source style vectors; we call this ``targeted restyling''. When combined with a novel ``tunable inference'' technique for controlling token add/delete rates, this gives our final model: \textbf{Text} \textbf{S}tyle \textbf{E}xtraction and \textbf{T}unable \textbf{T}argeted \textbf{R}estyling (TextSETTR).

\begin{figure*}
    \centering
    \includegraphics[width=0.88\linewidth]{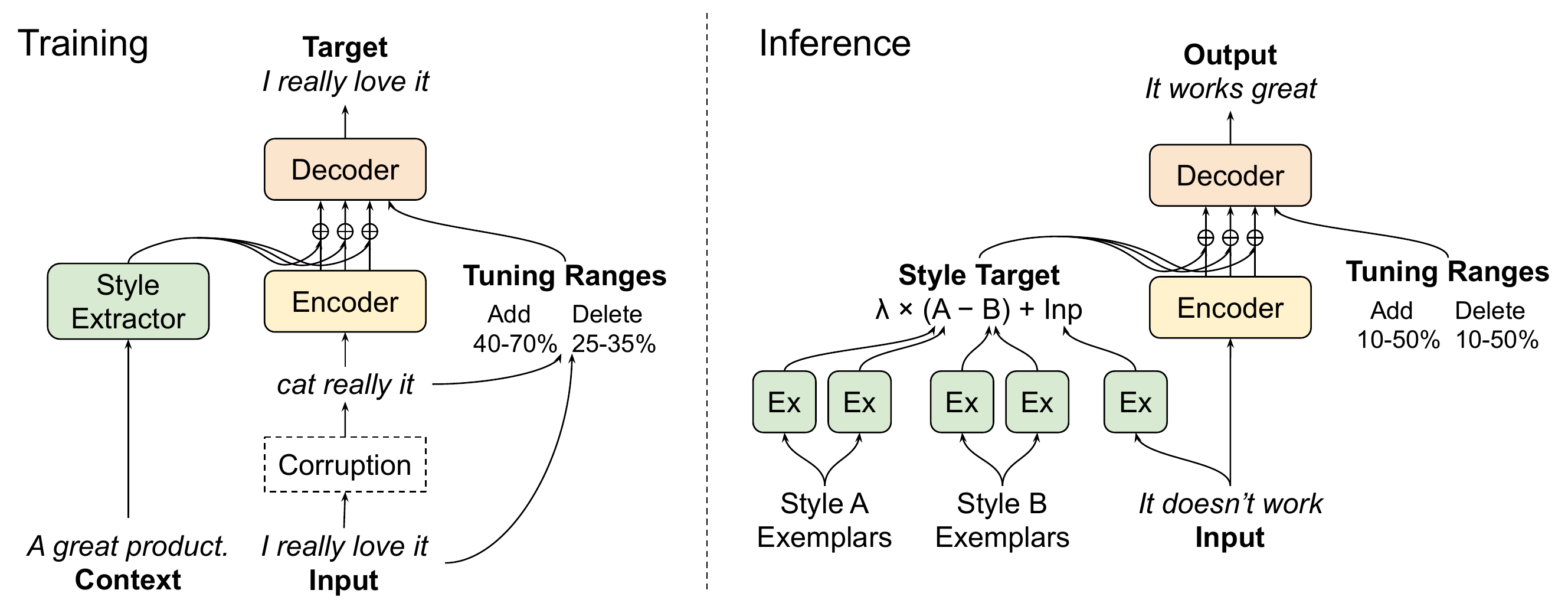}
    \caption{TextSETTR architecture for few-shot style transfer. The Encoder, Decoder and Style Extractor (Ex) are transformer stacks initialized from pretrained T5\@. During training, the model reconstructs a corrupted input, conditioned on a fixed-width ``style vector'' extracted from the preceding sentence. At inference time, a new style vector is formed via ``targeted restyling'': adding a directional delta to the extracted style of the input text. Stochastic tuning ranges provide extra conditioning for the decoder, and enable fine-grained control of inference.}
    \label{fig:architecture}
\end{figure*}

Our main contributions are to: (1) present a new, flexible approach to few-shot style transfer, (2) use sentence adjacency as a means for inducing text style representations, (3) reframe style transfer as ``targeted restyling'' directional operations in style space, (4) introduce ``tunable inference'' for finer-grained control of transfers, (5) show the effectiveness of ``noisy'' back-translation training, and (6) illustrate few-shot generalization to a range of style attributes including dialect, emotiveness, formality, politeness, and sentiment.

\section{Method}\label{section:method}

Figure~\ref{fig:architecture} illustrates our proposed TextSETTR architecture.
At a high level, our approach follows \citet{lample19}, who train a denoising auto-encoder conditioned on a fixed-width style vector. The key difference in our case is that the true style is unknown at training time. To overcome this, we jointly train a ``style extractor'' component to induce a useful style representation (that can aid in reconstruction) from text in the nearby context. We describe this in more detail below.

\subsection{Model Architecture}

We conduct our experiments using a modified version of the Text-to-Text Transfer Transformer (T5) \citep{2020t5}.
Like T5, our model includes a transformer-based encoder and decoder. As in T5 pretraining, the input to the encoder is a corrupted version of the target, resulting in a reconstruction task. Our goal is to design a type of corruption that results in this training task resembling style transfer, despite the lack of labeled training data.

Our core addition to T5 is the style extractor. This component's architecture is based on that of the encoder, and its input is an uncorrupted sentence in the same style as the target; relying on our assumption that style is a slow-moving feature, we use the sentence preceding the target (the ``context'') for this. This encourages extracting a style representation that is useful for repairing the corrupted input. We note that this can result in a representation that encodes slow-moving attributes in general, which may include some features that do not fit an intuitive definition of textual style (such as topic).

The only architectural difference between the encoder and style extractor is that we mean-pool the style extractor's hidden state sequence into a single fixed-width ``style vector''; in our experiments, the dimensionality of this vector and the encoder hidden states is 1024. To incorporate the style vector into the rest of the model, we simply add it to each of the final encoder hidden states.

We initialize the weights of our model with those of a pretrained T5 model. We initialize both the style extractor and encoder from the pretrained encoder, but the weights are \textit{not} tied during training.

\subsection{Corruption Strategies}
We experiment with combinations of three different reconstruction tasks, each contributing a loss term. All three share the same overall structure, where a sentence $s_i$ in the dataset is corrupted by some function $f$ to produce $\tilde{s}_i = f(s_i)$. The cross-entropy loss is calculated using the uncorrupted sentence $s_i$ as the target, the corrupted sentence $\tilde{s}_i$ as the input, and the uncorrupted preceding sentence $s_{i-1}$ as the context. The three choices of $f$ are Noise (N), Back-Translation (BT), and Noisy Back-Translation (NBT), described below.

\textbf{Noise (N)}
This function corrupts the input by (i) dropping, (ii) replacing, and/or (iii) shuffling tokens, in that order. For each example we sample a separate noise probability $p$ for each sub-type of noise from a uniform distribution in the range 20--60\%; doing so should widen the model's range of possible style transfers at test time.

For \textit{drop} noise, we drop each token in $s_i$ with probability $p$. For \textit{replace} noise, let $s_{ik}$ be the $k$-th token within $s_i$. For each $s_i$, a random other example $s_j$ is chosen, and then each token $s_{ik}$ is replaced by $s_{jk}$ with probability $p$. If $s_j$ has fewer than $k$ tokens, then the replacement does not occur. For \textit{shuffle} noise, each token in $s_i$ is chosen with probability $p$, and then all chosen tokens are randomly shuffled to the position of another chosen token, leaving non-chosen tokens in place.

The use of drop and shuffle noise results in a loss term similar to the denoising loss used by \citet{lample19}. Their motivation for this loss was to encourage language modeling. As we fine-tune an already-strong T5 language model in our experiments, our motivation is rather to introduce a \textit{conditional} element to the language model, in the form of the extracted style vector input.

\textbf{Back-Translation (BT)}
This corruption function, used by \citet{lample19}, runs the current version of the model in inference mode to transfer $s_i$ into a different style, giving the corrupted $\tilde{s}_i$. In prior work using labels, specifying a different target style was straightforward. In our case, because we do not have access to labels, we simply sample a random sentence $s_j$ to use as the context. To increase diversity of the generated examples, we decode with sampling instead of greedy decoding.

Because $\tilde{s}_i$ is produced by a strong language model, BT should result in examples where both the input and output are coherent sentences, matching our inference setting. By contrast, Noise corruption does not resemble test-time inputs.

\textbf{Noisy Back-Translation (NBT)}
This novel corruption function is a composition of the previous two. Noise is first applied to $s_i$ as described above, and the result is used as the input (with randomly-sampled $s_j$ as the context) to the model in inference mode to produce $\tilde{s}_i$ via sampling, as in BT.

Once the model has learned to undo random noise, NBT should produce training examples where some of the tokens are preserved from $s_i$ while others were generated by the model itself under the influence of the ``incorrect'' context $s_j$. This is similar to BT, but we hypothesize that it may be better suited to style transfer. BT was originally used for machine translation \citep{sennrich16}, a setting where most or all input tokens need to change. In contrast, style transfer within a single language usually requires only changing a subset of tokens; the training examples resulting from NBT should have this property. We believe that this will encourage the model to identify which tokens in the input do not match the target style indicated by $s_{i-1}$ and change them, which is exactly what we want a style transfer model to do.

\textbf{Final Loss}
The final loss term used for training is the sum of the above loss terms, each calculated from the same input $s_i$. However, not every model we experiment with includes all three losses.

\subsection{Inference Procedure}\label{section:inference}

\textbf{Tunable Add/Delete Rates}
In preliminary experiments, we observed a recurring problem that the model would often change either far too little (failing to achieve the target style), or far too much (failing to preserve the input content). To address this problem, we introduce a ``tunable inference'' mechanism to constrain how much content should be added and deleted at inference time.

For every input/output pair during training, we calculate the proportions of tokens that were added and deleted. The ``add rate'' is the proportion of output tokens absent from the input, and the ``delete rate'' is the proportion of input tokens absent from the output.\footnote{This calculation ignores word order. As one example, if a token appears three times in the input and five times in the output, two of the five occurrences are counted as ``added''.} We provide these rates to the decoder as \textit{ranges} covering but not necessarily centered on the true rates.\footnote{Specifically, we sample each range width uniformly from [0,1], and uniformly sample the ``alignment'' of the true rate within the range. The final ranges are clipped to [0,1], and a vector containing the upper and lower bound of each range is prepended to the encoder hidden state sequence.} This approach provides more flexibility at inference time, so we can enforce tight or loose constraints on each rate.

\textbf{Targeted Restyling}
While previous work on style transfer has largely assumed a fixed set of discrete styles, we expect our model's learned style representations to capture a rich summary of the sentence covering many attributes without specifying them beforehand.
For example, a given style vector might encode that a sentence is informal, humorous, in British English, and so on.

In this framework, transferring a single attribute (e.g.,~informal\,$\rightarrow$\,formal) is not as simple as just providing a vanilla ``formal'' style target, as this would ignore all the other attributes that defined the original input. Rather, we must operate in style space to construct a new target style that is simultaneously formal, humorous, British, and so on.

Concretely, at inference time, we assume access to a small set of ``exemplar'' sentences (between 1 and 100) for both the source value (e.g.,~informal) and target value (e.g.,~formal) of the attribute being modified. We infer style vectors for each exemplar using the style extractor, and take the mean of each class, giving vectors $v^{src}$ and $v^{trg}$. Assuming the exemplar pools are relatively diverse, this averaging should ``wash out'' most untargeted attributes.

To transfer an input sentence $x$, we apply a targeted restyling in the appropriate direction. After extracting the original style from the input itself, $v^x$, we compute the target output style by moving in the direction of the delta between the source and target attributes values, as in (\ref{delta_formula}), producing the style vector used for decoding. In practice, we find that the delta scale $\lambda$ is an important hyperparameter to tune. Generally values in the range [1.0,\,10.0] work well, with the best values depending on the attribute and the exemplars in question.

\vspace{-0.5em}
\begin{equation}\label{delta_formula}
v^x + \lambda \times \left ( v^{trg} - v^{src} \right )
\end{equation}

\begin{figure*}
  \begin{minipage}[c]{0.5\textwidth}
    \setlength{\tabcolsep}{0.25em}
    \resizebox{0.85\columnwidth}{!}{
    \footnotesize
    \begin{tabular}{lcccc}
    \toprule
    & \textbf{Model} & \textbf{Acc.} & \textbf{Content} & \textbf{G} \\
    \midrule
    \multirow{14}{*}{\textbf{Few-Shot}}
    & TextSETTR (10--30\%) & 54.0 & 55.8 & \textbf{54.9} \\
    & TextSETTR (20--40\%) & 73.7 & 34.7 & 50.6 \\
    & N & 23.4 & 84.4 & 44.4 \\
    & NBT & 70.0 & 27.8 & 44.1 \\
    & N\,$+$\,BT & 13.3 & 98.7 & 36.2 \\
    & $-$replace noise & 66.1 & 42.1 & 52.8 \\
    & $+$shuffle noise & 70.3 & 34.1 & 49.0 \\
    & manual exemplars & 52.4 & 44.2 & 48.1 \\
    & 1000 exemplars & 74.5 & 37.2 & 52.6 \\
    & $-$tunable inference & 71.5 & 39.4 & 53.1 \\
    & overwrite style & 25.3 & 55.8 & 37.6 \\
    & small train set & 74.5 & 33.4 & 49.9 \\
    \cmidrule{2-5}
    & CP-G & 51.1 & 35.5 & 42.6 \\
    & CP-B & 36.3 & 39.8 & 38.0 \\
    \midrule
    \multirow{3}{*}{\textbf{Labeled}}
    & CrossAligned & 68.2 & 2.9 & 14.1 \\
    & Delete\&Retrieve & 49.4 & 56.9 & 53.0 \\
    & B-GST & 60.2 & 54.2 & \textbf{57.1} \\    
    \bottomrule
    \end{tabular}}

  \end{minipage}
  \begin{minipage}[c]{0.50\textwidth}
    \includegraphics[trim=0 16 41 41,clip,scale=0.57, right]{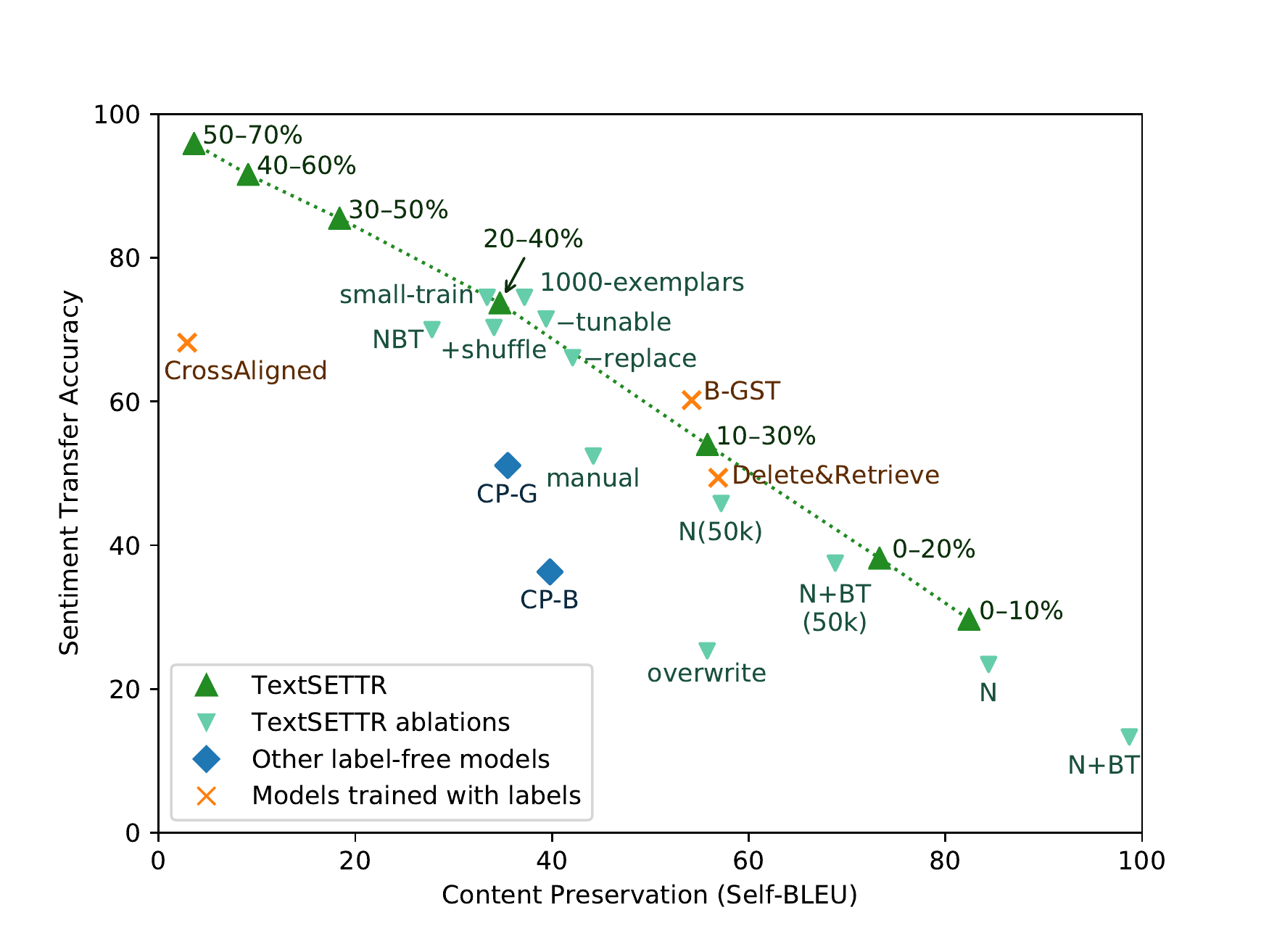}
  \end{minipage}
  \caption{Automatic evaluation metrics comparing our TextSETTR model, ablations, and previous work. Up-and-right is better. We train for 10k steps and use add/delete:20--40\% unless otherwise specified. We recalculate metrics for previous approaches, using our BERT classifier for accuracy, ensuring direct comparability.}\label{fig:results}
\end{figure*}

\section{Experiments on Sentiment Transfer}
\label{section:sentiment}

To evaluate our approach and better understand the effects of our various design choices, we test on few-shot sentiment transfer, using the Amazon reviews dataset of \citet{li18}. However, as their training split doesn't indicate which sentences were adjacent in the original reviews, we make use of a different source of raw review text.

\textbf{Training Procedure} Our unlabeled training data comes from the 233.1M Amazon reviews provided by \citet{ni-etal-2019-justifying}. Ignoring the star ratings completely, we extract adjacent lines from multi-line reviews to use as the context and input for our training procedure, giving 23.6M examples. We also preprocess all text to match the format of the \citet{li18} data, as detailed in Appendix~\ref{sec:appendix_preprocessing}.
Initializing our model from pretrained T5 (t5.1.1.large), we fine-tune on these examples, optimizing the joint reconstruction loss from Section~\ref{section:method}. Our default TextSETTR configuration is selected based on preliminary experiments (on development data) varying  the set of reconstruction tasks and inference procedures. The model uses an equally weighted combination of the Noise (N) and Noisy Back-Translation (NBT) tasks. For both tasks, we use drop and replace noise, but no shuffle noise. We fine-tune for 10k steps, with a batch size of 65,536 tokens, and a fixed learning rate of 1e-3. 

\textbf{Evaluation Procedure} Following prior work, we use automatic metrics to assess attribute control (sentiment) and content preservation on the data from \citet{li18}. To estimate the sentiment of the output, we fine-tune a BERT-Large classifier \citep{bert} on the train split, scoring 87.8\% accuracy on the dev split. For content preservation, we follow \citet{sudhakar19} and \citet{xu20} and calculate self-BLEU between the output and input, using SacreBLEU \citep{post18}.\footnote{Version string: BLEU+case.mixed+numrefs.1+ smooth.exp+tok.13a+version.1.4.13}\textsuperscript{,}\footnote{Some prior work reports instead BLEU scores between outputs and human-generated transfers from \citet{li18}; we found this to be highly correlated with self-BLEU but report it in Appendix~\ref{sec:reference_bleu} for completeness.} Following \citet{xu-etal-2018-unpaired}, we report ``G-score'' (the geometric mean of accuracy and content) as a summary of overall model quality.

To perform transfers, we follow the procedure from Section \ref{section:inference}. For our default setup, we sample 100 positive and 100 negative exemplars from the \citet{li18} train split. Unless otherwise specified, we use greedy decoding, a delta scale of $\lambda$=8, and add/delete tuning ranges of 20--40\%.

\textbf{Core Results} Figure~\ref{fig:results} shows our core results. Our default TextSETTR configuration (N+NBT training, tuning ranges 20--40\%) achieves 73.7\% classifier-judged accuracy at swapping sentiment, while still staying somewhat close to the original input text (self-BLEU 34.7). Due to our tunable inference technique, we can also trade off accuracy for content preservation by adjusting the add/delete rates, as seen in the points along the green line. Notably, TextSETTR outperforms the few-shot CP-G and CP-B models of \citet{xu20}. More remarkably, TextSETTR outperforms several approaches that rely on training labels: CrossAligned \citep{shen17} and Delete\&Retrieve \citep{li18}. However there is still a small gap between our few-shot approach and the best labeled model, B-GST \citep{sudhakar19}.

\begin{table}
\centering
\resizebox{0.85\columnwidth}{!}{
\footnotesize
\begin{tabular}{cccc}
\toprule
\textbf{Model} & \textbf{Accuracy} & \textbf{Content} & \textbf{G} \\
\midrule
TextSETTR (10--30\%) & 72.7 & 60.2 & 66.2 \\
TextSETTR (20--40\%) & 83.6 & 39.4 & 57.4 \\
\midrule
\citealt{lample19} & 82.6 & 54.8 & \textbf{67.3} \\
\bottomrule
\end{tabular}}
\caption{Comparison with \citet{lample19} on the setting that includes pos$\rightarrow$pos and neg$\rightarrow$neg transfers.}\label{tab:results_easy}
\end{table}

In Table~\ref{tab:results_easy}, we compare with \citet{lample19} on the evaluation setting including pos$\rightarrow$pos and neg$\rightarrow$neg transfers. This setting doesn't match our inference procedure, which assumes that the input and output styles differ. Nevertheless, TextSETTR comes close to the performance of \citet{lample19}, despite not benefiting from training labels.

\begin{table}
\centering
\resizebox{0.98\columnwidth}{!}{
\footnotesize
\begin{tabular}{cccc}
\toprule
\textbf{Model} & \textbf{Sentiment} & \textbf{Preservation} & \textbf{Fluency} \\
\midrule
TextSETTR (10--30\%) & 2.0 & 3.5 & 2.9 \\
TextSETTR (20--40\%) & 2.5 & 2.6 & 4.0 \\
\midrule
Delete\&Retrieve     & 2.5 & 3.1 & 3.3 \\
B-GST                & 2.2 & 2.9 & 3.6 \\
\bottomrule
\end{tabular}}
\caption{Human evaluation metrics.}\label{tab:human}
\end{table}

As automatic metrics can diverge from human judgment \citep{sudhakar19}, we also conduct human evaluations of the three strongest models from Figure~\ref{fig:results}. We sample 200 examples per transfer direction from the \citet{li18} test set, and ask three annotators to evaluate each input/output pair on three metrics: sentiment transfer (how well the model changed the sentiment), content preservation, and fluency, on scales of 1--5. The results in Table \ref{tab:human} confirm that TextSETTR achieves similar quality to models that benefit from training labels. Further details are presented in Appendix~\ref{sec:appendix_human}.

\subsection{Ablations}

\textbf{Modifying Inference Procedure} To better understand the value of our proposed ``targeted restyling'' mechanism, we consider an alternative inference procedure where we ignore the style of the input and simply use the average target exemplar style $v^{trg}$ as the style vector. We expect that since our learned style space covers multiple attributes, this will result in setting the target attribute (e.g.~sentiment) while simultaneously \emph{overwriting} all other style attributes (e.g.~formality) using the average style of the target exemplars. This is borne out in our ``overwrite style'' ablation, which performs significantly worse than our baseline: accuracy drops from 54.0\% to 25.3\% with no gain in self-BLEU.

To assess the value of tunable add/delete rates, we also train a model ($-$tunable) without this feature. While the automatic metrics are slightly above the TextSETTR line, we observe several advantages to the tunable model. For one, we observe it significantly reduces the variance in self-BLEU across different inputs. For example, focusing on the case of overly high self-BLEU, we find that without tunable inference, 14.6\% of dev eval outputs are identical to their inputs, whereas with tunable inference, this goes to 0.9\%. Additionally, through qualitative analysis in Section~\ref{section:one_model}, we find that tunable inference allows more flexibility for controlling different types of transfer.

\textbf{Adjusting Data Sizes} While our unlabeled training data set consists of 23.6M examples, our model only sees 5.1M of these over its 10k steps of training. Yet this is still nearly 10$\times$ more data than the 0.6M examples in the \citet{li18} training set used by previous approaches. For a more direct comparison, we experiment with a ``small train set'', sampling 0.6M examples from our training set. Remarkably, the results in Figure~\ref{fig:results} are nearly identical to our baseline, supporting our hypothesis that a fairly lightweight adaptation is sufficient to allow T5 to extract and transfer textual style.

To test the limits of our model's generalization, we reduce the set of exemplars to four manually selected examples of each class. In this setting, we also find reducing delta scale to $\lambda$=4 is beneficial. The results, shown as ``manual exemplars'' in Figure~\ref{fig:results}, are still competitive, indicating that our approach generalizes well to this very-few-shot inference setting. In the other direction, we find that increasing the number of sampled exemplars from 100 to 1000 only provides small additional gains.

\textbf{Modifying Training Task} \citet{lample19} showed promising results by combining noise (N) with back-translation (BT). However we find this combination unstable.\footnote{For all experiments in the paper, we use 0.0 for the add/delete rates during the forward pass of back-translation. However we later found that using \emph{random} add/delete rates in back-translation can improve performance in the N+BT setting. On sentiment transfer, this improved our N+BT ablation to self-BLEU 42.4, accuracy 71.4\%, G-score 55.0.} When training for 10k steps, our N and N+BT models nearly always copy their input. Training for 50k steps recovers reasonable performance, but the metrics still fall below the TextSETTR line, using our novel NBT task. We also experiment with using NBT in isolation, but this again underperforms our baseline. We expect that the denoising task helps to ensure the NBT inputs (themselves the outputs of denoising) consist of realistic well-formed text. Finally, while \citet{lample19} use drop and shuffle noise, we find that only drop and replace are valuable.

\addtolength{\tabcolsep}{-3pt}

\begin{figure*}
    \centering
    \footnotesize
    \begin{tabular}{m{2.3cm} >{\centering\arraybackslash}m{4.95cm} >{\centering\arraybackslash}m{4.95cm}}
    \textbf{Before} TextSETTR training (pretrained T5 initialization) & 
    \includegraphics[scale=0.31,trim=15 16 15 16,clip]{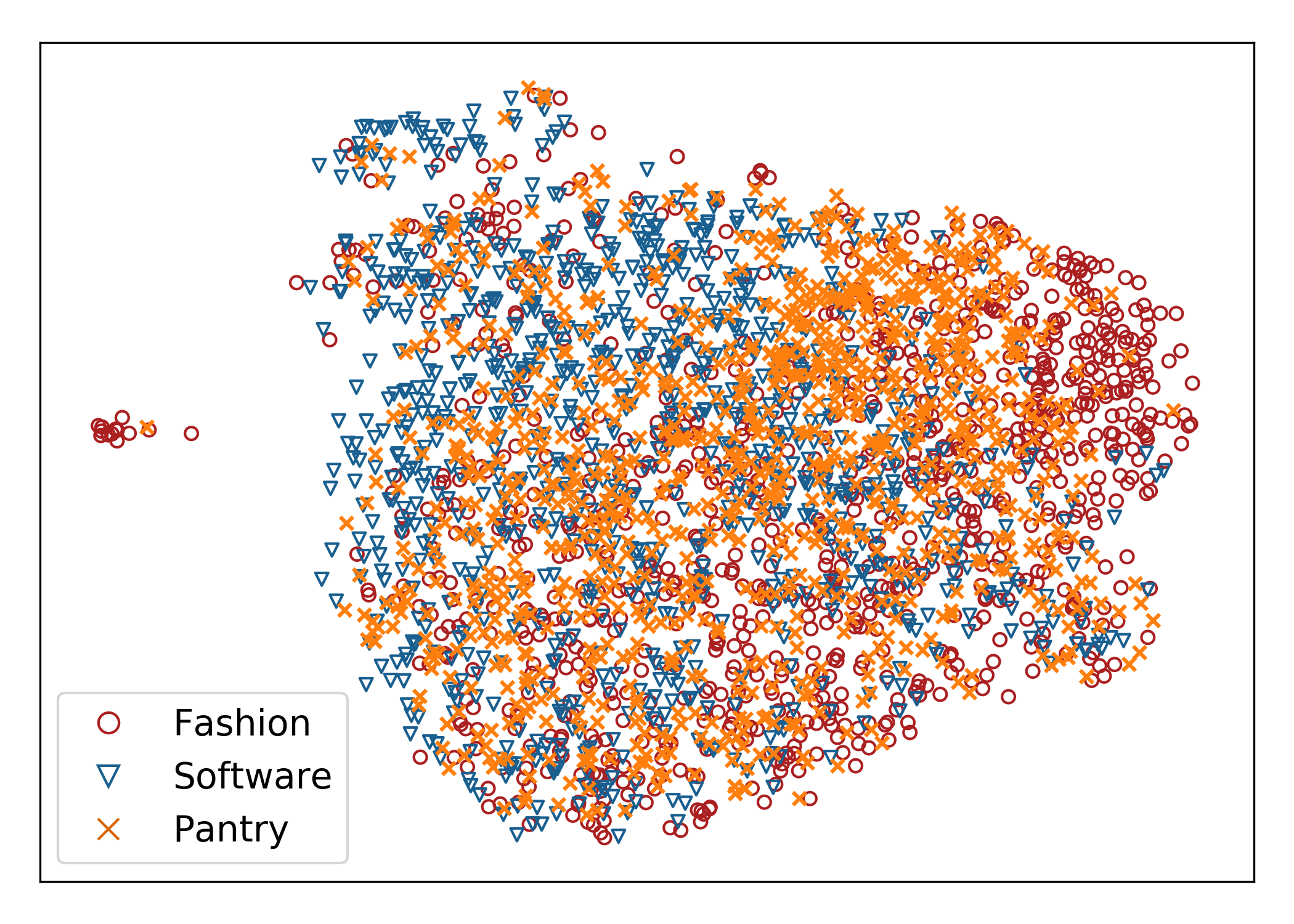} &
    \includegraphics[scale=0.31,trim=15 16 15 16,clip]{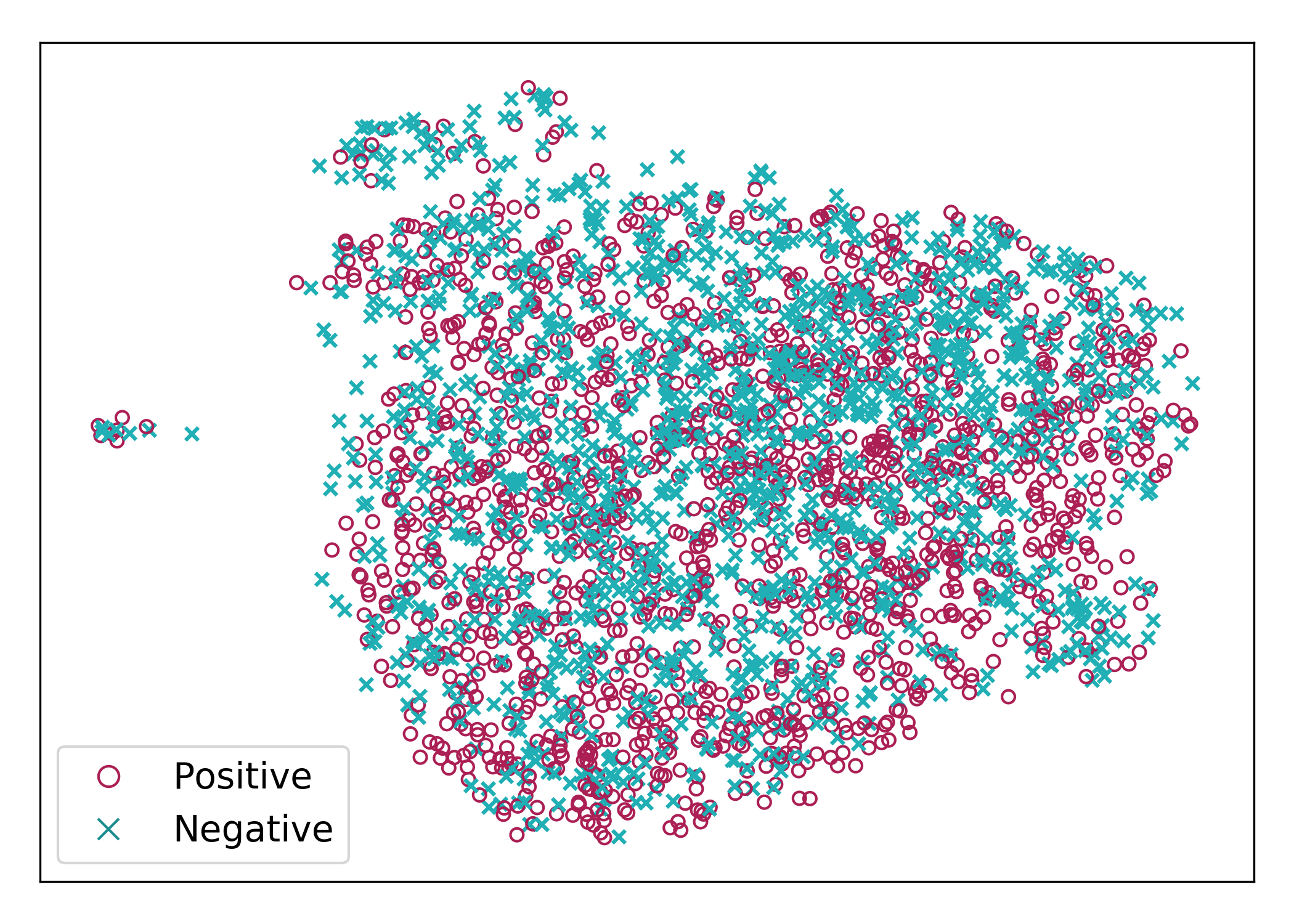} \\
    \textbf{After} TextSETTR training & \includegraphics[scale=0.31,trim=15 16 15 16,clip]{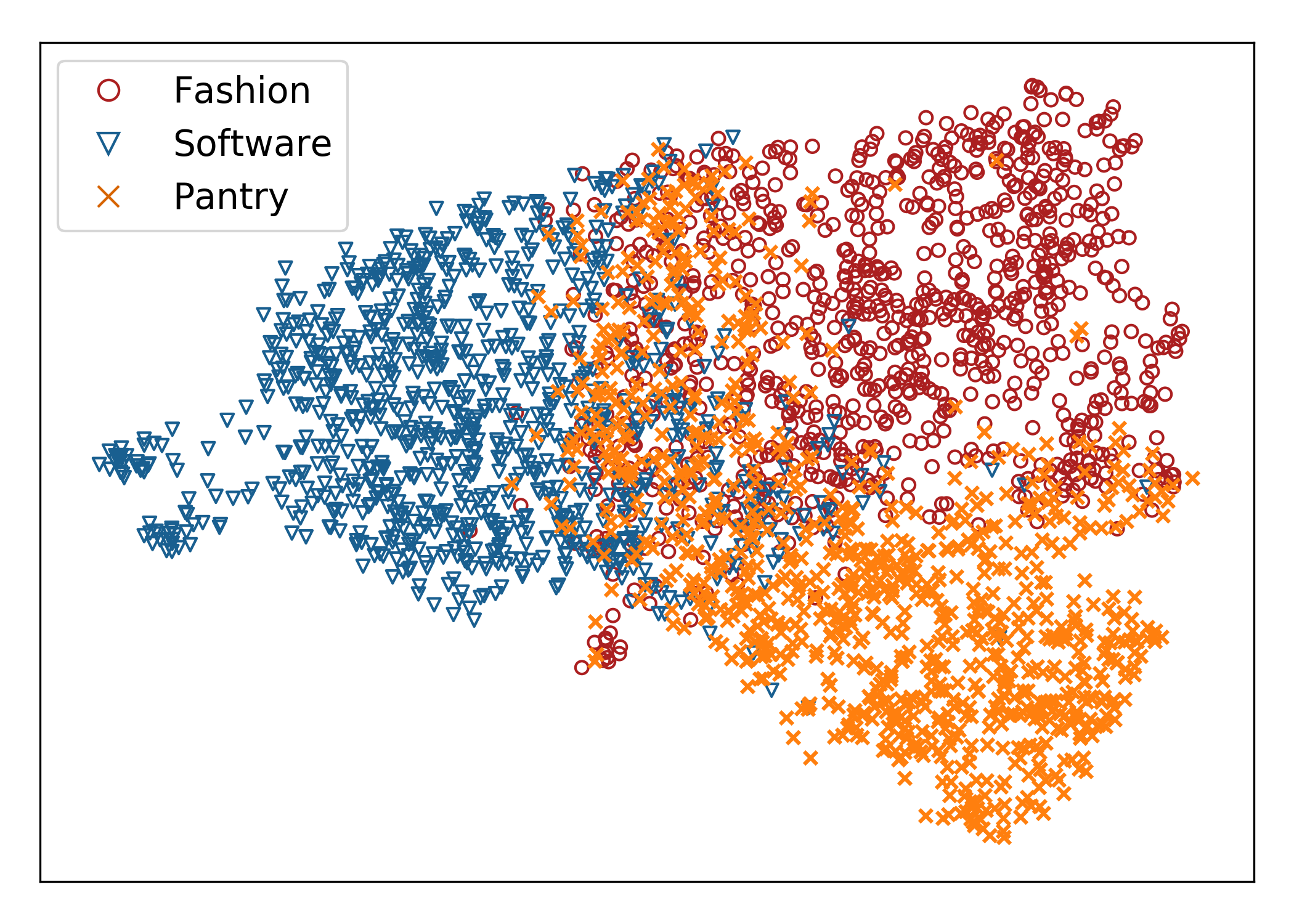} &
    \includegraphics[scale=0.31,trim=15 16 15 16,clip]{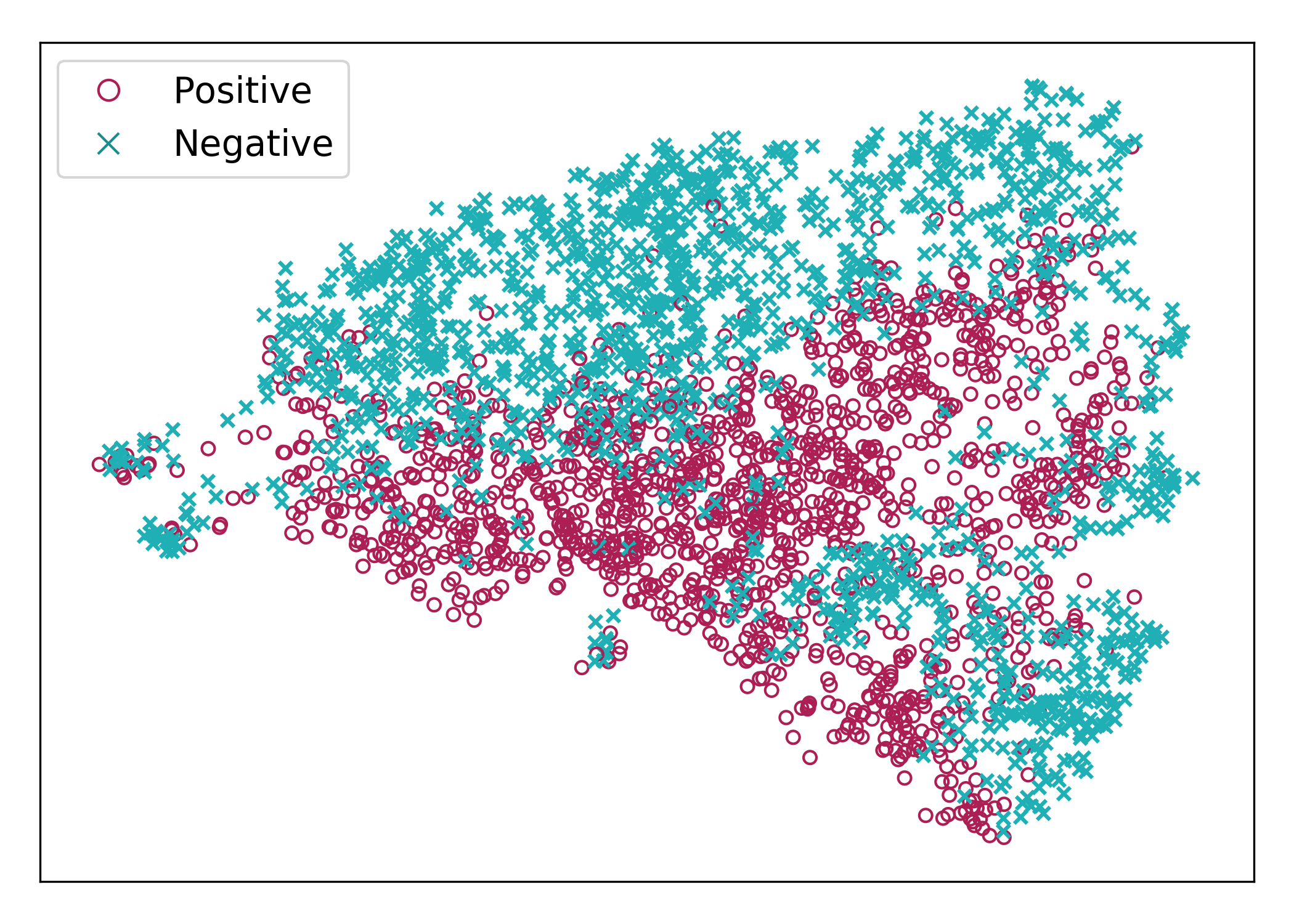} \\
    \end{tabular}

    \caption{2D UMAP embeddings of the style vectors extracted by our TextSETTR model before and after training, for text inputs from Amazon reviews covering three product categories and two sentiment labels. Within each row, the same embeddings are labeled with product category (left) and sentiment (right). We sub-sample to 3,000 points after dimensionality reduction. Note, we don't expect perfect separation, as inputs may be underspecified for category (``I love this product'') or for sentiment (``I bought this last month''). We also don't expect to see crisp linear separation within each attribute since we aim for the learned embedding space to encode many style attributes simultaneously.}
    \label{fig:umap}
\end{figure*}

\addtolength{\tabcolsep}{3pt}

\newcommand{\mycell}[1]{\makecell[{{>{\hangindent1.47em}p{6.5cm}}}]{#1}}

\begin{table*}

\fontsize{8}{9.9}\selectfont
\setlength{\tabcolsep}{0.5em}
\centering

\begin{tabular}{p{0.47\textwidth}|p{0.47\textwidth}} \toprule

\textbf{Reserved $\Rightarrow$ Emotive} & \textbf{Emotive $\Rightarrow$ Reserved} \\ \hline
\mycell{I \red{\underline{liked the}} movie\red{\underline{.}}\\
$\Rightarrow$ I \blue{\underline{cannot even describe how amazing this}} movie \blue{\underline{was!!}}}
 & 
\mycell{I \red{\underline{loved every minute of}} the movie\red{\underline{!}}\\
$\Rightarrow$ I \blue{\underline{liked}} the movie\blue{\underline{.}}} \\ \hline

\mycell{I was \red{\underline{impressed}} with the results\red{\underline{.}}\\
$\Rightarrow$ I was \blue{\underline{absolutely blown away}} with the results\blue{\underline{!!}}}
 & 
\mycell{I was \red{\underline{shocked}} by the \red{\underline{amazing}} results\red{\underline{!}}\\
$\Rightarrow$ I was \blue{\underline{surprised}} by the results\blue{\underline{.}}} \\

\midrule
\textbf{American $\Rightarrow$ British} & \textbf{British $\Rightarrow$ American} \\ \hline

\mycell{The \red{\underline{elevator}} in my \red{\underline{apartment}} isn't working.\\
$\Rightarrow$ The \blue{\underline{lift}} in my \blue{\underline{flat}} isn't working.}
&
\mycell{The \red{\underline{lift}} in my \red{\underline{flat}} isn't working.\\
$\Rightarrow$ The \blue{\underline{elevator}} in my \blue{\underline{apartment}} isn't working.} \\
\hline

\mycell{The \red{\underline{senators}} will return to \red{\underline{Washington}} next week.\\
$\Rightarrow$ The \blue{\underline{MPs}} will return to \blue{\underline{Westminster}} next week.}
 & 
\mycell{\red{\underline{MPs}} will return to \red{\underline{Westminster}} next week.\\
$\Rightarrow$ \blue{\underline{Representatives}} will return to \blue{\underline{Washington}} next week.} \\

\midrule
\textbf{Polite $\Rightarrow$ Rude} & \textbf{Rude $\Rightarrow$ Polite} \\ \hline

\mycell{\red{\underline{Are you positive}} you've understood my point\red{\underline{?}}\\
$\Rightarrow$ you've \blue{\underline{never}} understood my point\blue{\underline{!}}} 
&
\mycell{\red{\underline{What}} the \red{\underline{hell}} is \red{\underline{wrong}} with your attitude\red{\underline{?}}\\
$\Rightarrow$ \blue{\underline{Perhaps}} the \blue{\underline{question}} is \blue{\underline{more about}} your attitude\blue{\underline{.}}} \\
\hline

\mycell{\red{\underline{Could}} you ask \red{\underline{before}} using my phone\red{\underline{?}}\\
$\Rightarrow$ \blue{\underline{I}} ask you \blue{\underline{to stop}} using my phone\blue{\underline{!}}}
 & 
\mycell{I could \red{\underline{care less}}, \red{\underline{go}} find somebody else to do this \red{\underline{crap}}.\\
$\Rightarrow$ I could \blue{\underline{be wrong}}, \blue{\underline{but I would try to}} find somebody else to do this.} \\

\midrule
\textbf{Formal $\Rightarrow$ Informal} & \textbf{Informal $\Rightarrow$ Formal} \\ \hline

\mycell{\red{\underline{I hereby commit to}} never \red{\underline{purchase}} anything from this \red{\underline{institution in the future}}.\\
$\Rightarrow$ \blue{\underline{i gonna}} never \blue{\underline{buy}} anything from this \blue{\underline{place again}}.}
&
\mycell{\red{\underline{best}} book \red{\underline{ever!!}}\\
$\Rightarrow$ \blue{\underline{The}} book \blue{\underline{is highly recommended.}}} \\
\hline

\mycell{\red{\underline{I couldn't figure out}} what \red{\underline{the author was}} trying to say.\\
$\Rightarrow$ \blue{\underline{i dont know}} what \blue{\underline{ur}} trying to say.}
 & 
\mycell{\red{\underline{couldnt}} figure out what author \red{\underline{tryna}} say\\
$\Rightarrow$ \blue{\underline{The reader couldn't}} figure out what \blue{\underline{the}} author \blue{\underline{was}} \blue{\underline{trying to}} say.} \\

\midrule
\textbf{Positive $\Rightarrow$ Negative} & \textbf{Negative $\Rightarrow$ Positive} \\ \hline

\mycell{I was pretty \red{\underline{impressed}} with the results.\\
$\Rightarrow$ I was pretty \blue{\underline{disappointed}} with the results.}
&
\mycell{I was pretty \red{\underline{disappointed}} with the results.\\
$\Rightarrow$ I was pretty \blue{\underline{impressed}} with the results.} \\
\hline

\mycell{I will definitely buy this brand again.\\
$\Rightarrow$ I will definitely \blue{\underline{not}} buy this brand again.}
 & 
\mycell{I definitely won't buy this brand again.\\
$\Rightarrow$ I definitely won't \blue{\underline{hesitate to}} buy this brand again.} \\

\bottomrule
\end{tabular}
\caption{Examples of transferring along five different axes of style. The same model is used across all examples, with no additional training. Words deleted from the input are \red{\underline{red}}, and words added in the output are \blue{\underline{blue}}. Within each category, a fixed tiny set of exemplars is chosen, and fixed delta scale and tuning rates are used. The exemplars and settings are provided in Appendix~\ref{sec:appendix_settings}.}
\label{tab:cherrypicks}

\end{table*}

\subsection{Embedding Visualization}

To demonstrate that our learned style extractor encodes multiple aspects of textual style, we compute style vectors for 12,000 lines of text from three review categories (Fashion, Software, Pantry) from the \citet{ni-etal-2019-justifying} Amazon data. Within each category, we sample 2,000 positives (4 or 5 star) and 2,000 negatives (1 or 2 star), filtering examples where our BERT classifier disagrees with the label. Figure~\ref{fig:umap} (bottom) plots a 2D UMAP dimensionality reduction \citep{umap} of the vectors, and shows clear separations among sentiments and product categories. The top row runs UMAP with the same settings, but over style vectors from our model \emph{before} training, where the style extractor is initialized from pretrained T5. The contrast is a clear indication that our training procedure is helping to learn a representation space where sentiment and topic values are well separated.

To confirm that the observed separation isn't an artifact of dimensionality reduction, we compute the average distance between style vectors (a) \emph{within} a class, and (b) \emph{across} classes. We measure ``separation'' as the relative increase in mean distance between these two conditions. For product category, we find TextSETTR training improves separation from $1.7$\% to $8.1$\%. For sentiment, TextSETTR training improves separation from $0.9$\% to $4.7$\%.

\section{One Model for All Styles}\label{section:one_model}

An advantage of few-shot style transfer is that, in theory, a single model can perform transfer along any ``dimension'' of style given only a few exemplars, without the need for additional training. In this section, we investigate the degree to which our approach achieves this goal in practice. For this purpose, we train a single general-purpose TextSETTR model, with the same configuration as our model from Section~\ref{section:sentiment}, except fine-tuned for 200k steps on English Common Crawl data (the same ``C4'' data that T5 pretrained on) instead of Amazon reviews.

\textbf{Qualitative Evaluation} Given that our architecture limits the style representation to 1024 dimensions, one may ask how the unsupervised model will make use of this capacity, and which style attributes will be encoded in the learned space. Encouragingly, we find that our model trained on unlabeled Common Crawl data is capable of transferring along many independent axes of style. Table~\ref{tab:cherrypicks} shows selected successful examples of our Common Crawl model transferring emotiveness, dialect, politeness, formality and sentiment. The same model is used in each case, with no additional training. At inference time, a tiny set of exemplars (1--5 examples of each class) is the only labeled data used to compute the style vector delta; these exemplars are presented in Appendix~\ref{sec:appendix_settings}.

Across each type of transfer, we see evidence of generalization beyond the specifics of the chosen exemplars. In making text more emotive, the model uses \textit{amazing} and \textit{blown away}, despite these terms not occurring in the exemplars. In making text more polite, the model inserts novel hedges like \textit{perhaps} and \textit{I could be wrong}. In transferring between American and British styles, the model generalizes to unseen vocabulary items (\textit{elevator}\,$\leftrightarrow$\,\textit{lift}) and draws sound analogies (\textit{senators}\,$\leftrightarrow$\,\textit{MPs}). We do note though that the latter case illustrates that the model is willing to change the semantic content of the input in cases where it would otherwise be out-of-place in the target style. Future work includes investigating ways to control this in settings where such behavior is not desired.

\textbf{Quantitative Evaluation} To assess the quality of our general-purpose TextSETTR model, we benchmark the same model on three distinct transfer tasks in Table~\ref{tab:c4_results}.\footnote{For each task, we set our tuning ranges to 20--40\% and compute target styles using 100 exemplars of each class taken from the train set. We use $\lambda$ values of sentiment:8, author:16, personality:8. To measure accuracy, we fine-tune BERT-Large classifiers over the training data, reaching validation accuracies of sentiment:87.8\%, author:89.7\%, personality:81.9\%.} The \textbf{sentiment transfer} task follows the evaluation procedure from Section~\ref{section:sentiment}. While our generic model underperforms our model trained on Amazon reviews, it still outperforms other few-shot methods. For \textbf{author transfer}, we use the Shakespeare-to-modern task of \citet{jhamtani17}. Here, TextSETTR outperforms the previous best model of \citet{he20} that leveraged 36,790 labeled examples during training. For \textbf{personality transfer}, we use the task of \citet{li-etal-2020-complementary}, which requires transferring between three personalities: angry, happy, malicious. We compare\footnote{Note, as \citet{li-etal-2020-complementary} use a different classifier to assess accuracy, those numbers may not be directly comparable.} TextSETTR, which sees no labels in training and only 100 of each class in inference, with CARA \cite{li-etal-2020-complementary}, which trained on 2,604 labels.

\begin{table}
\centering
\resizebox{0.92\columnwidth}{!}{
\footnotesize
\begin{tabular}{lcccc}
\toprule
\textbf{Task} & \textbf{Model} & \textbf{Acc.} & \textbf{Content} & \textbf{G} \\
\midrule
\multirow{6}{*}{\textbf{Sentiment}}
& CP-G & 51.1 & 35.5 & 42.6 \\
& CP-B & 36.3 & 39.8 & 38.0 \\
& TextSETTR & 44.9 & 54.4 & \textbf{49.4} \\
\cmidrule{2-5}
& CrossAligned & 68.2 & 2.9 & 14.1 \\
& Delete\&Retrieve & 49.4 & 56.9 & 53.0 \\
& B-GST & 60.2 & 54.2 & \textbf{57.1} \\
\midrule
\multirow{4}{*}{\textbf{Author}}
& UNMT & 68.5 & 7.8 & 23.1 \\
& BT+NLL & 59.3 & 12.4 & 27.1 \\
& \citealt{he20} & 68.5 & 12.5 & 29.2 \\
& TextSETTR & 81.7 & 13.8 & \textbf{33.5} \\
\midrule
\multirow{5}{*}{\textbf{Personality}}
& CARA & 91.6 & 21.6 & 44.5 \\
& CARA\textsubscript{AB} & 66.2 & 29.7 & 44.3 \\
& Ctrl-Gen & 67.6 & 22.9 & 39.3 \\
& ARAE$-$ & 88.0 & 20.3 & 42.3 \\
& TextSETTR & 49.3 & 46.0 & \textbf{47.6} \\
\bottomrule
\end{tabular}}
\caption{Automated metrics comparing our general-purpose TextSETTR model with recent work on three transfer tasks. To enable direct comparison, ``content'' refers to reference-BLEU for author transfer, and self-BLEU elsewhere. Apart from CP-G/CP-B, all competitors are trained for only one type of transfer using labeled data. Personality transfer results are from \citet{li-etal-2020-complementary}, while all others are recalculated from scratch.}\label{tab:c4_results}
\end{table}

\subsection{Dialect-Sensitive Completion}\label{sec:dialect_completion}

\begin{table*}[ht]

\fontsize{8}{9.9}\selectfont
\setlength{\tabcolsep}{0.5em}
\centering
\begin{tabular}{p{0.46\textwidth}|p{0.48\textwidth}} \toprule

\textbf{American $\Rightarrow$ British} & \textbf{British $\Rightarrow$ American} \\ \hline

My favourite food: fish and chips. &
My favorite food: quinoa.\\

My favourite hot drink: a mug of tea. &
My favorite hot drink: Starbucks Coffee.\\

My favourite dessert: a scone! &
My favorite dessert: a brownie.\\

My favourite city: Cardiff. &
My favorite city: San Diego. \\

My favourite band: The Beatles. &
My favorite band: The Black Keys.\\

My favourite sports league: the English Premier League. &
My favorite sports league: the NFL.\\

My favourite newspaper: The Daily Telegraph. &
My favorite newspaper: The Washington Post.\\

My favourite museum: the British Museum. &
My favorite museum: The National Air and Space Museum.\\

\bottomrule

\end{tabular}
\caption{Examples of dialect-sensitive completion ($\lambda$=8, add:40--70\%, delete:0\%). In each case, the input text consists of an unfinished phrase, for example: ``My favorite food: ''. The three exemplars used for each dialect are the same as those used for the transfers in Table~\ref{tab:cherrypicks}, as listed in Table~\ref{tab:dialect}.}
\label{tab:cherry_dialect_completion}
\end{table*}

In addition to performing style and attribute transfer, we find that our system can also be used as a style-aware language model capable of completing prompts in a specified style. Examples of completions in American and British English are given in Table~\ref{tab:cherry_dialect_completion}. In each case, the input is of the form ``My favorite X: ''. Despite the fact that TextSETTR is not trained specifically for completions, we can use the add/delete rates to encourage the model to insert a few additional tokens, while leaving the original prompt largely unchanged.\footnote{We note that in transferring American to British, the model prefers to change the prompt from \textit{favorite} to \textit{favourite}.}

The completions demonstrate knowledge of stereotypical American and British culture. It is remarkable that the model is able to generalize to ``deeper'' cultural differences such as music and drink preferences, given only the shallow vocabulary differences (e.g.,~\textit{neighbor} vs.~\textit{neighbour}) presented in the limited set of exemplars in Table~\ref{tab:dialect}.

It is also worth highlighting that, thanks to our directional transfer procedure, these completions are not merely ``typical American'' or ``typical British'' such as we would expect from a conditional language model trained on each sub-domain of text. Rather, since our inference procedure pushes the style away from one domain and towards the other, the resulting completions are \emph{distinctive} representations of each dialect. As one example, we expect ``quinoa'' would not only be a common American favorite, but also an uncommon British favorite.

Additional examples of using our model for tasks other than pure style transfer are presented in Appendix~\ref{sec:appendix_beyond}.

\section{Related Work}\label{section:related}

As mentioned at the outset, recent work on text style transfer falls into three classes: supervised, ``unsupervised'', and few-shot. Supervised style transfer has seen limited research due to the difficulty of obtaining parallel data. Examples include \citet{jhamtani17} and \citet{carlson18}.

\textbf{Unsupervised Approaches}
The bulk of research has focused on ``unsupervised'' approaches, which rely on labeled but non-parallel data.
Typically, labels are assumed to be available for both source and target styles (\citealt{shen17}, \citealt{li18}, \citealt{niu18}, and many others). \citet{zhao18} explore the case where only the target style is labeled. The use of labels at training time can aid modeling, but limits the applicability of these methods, as labeled datasets are not readily available for many attributes of interest.

Our work differs from the above by removing the need for training labels, and offering a single model that can target an unrestricted set of style attributes. Despite these differences, our work shares some similarities with past work. For example, our encoder-decoder architecture and corruption methods are similar to \citet{lample19}, and we leverage a strong pretrained language model, as in \citet{sudhakar19} and \citet{wu19}.

\textbf{Few-Shot Approaches}
A few-shot approach has recently been explored by \citet{xu20}.
The authors train a variational auto-encoder on unlabeled text, where a ``manipulable'' portion of the latent representation is constrained to fall on a k-dimensional simplex.
To perform transfer, they identify empirically the basis vector that most strongly corresponds to the target attribute, and manipulate its magnitude.
Compared to our approach, a key difference is that the number of latent factors must be chosen ahead of time, which limits the number of attributes that may be controlled. Additionally, there is no guarantee that a single basis of the learned simplex will correspond to a target attribute such as dialect or politeness.

\textbf{Controlled Generation}
A separate strand of research explores ``controlled generation'' methods for supplementing generative language models to allow control of specific attributes of the output text. As with style transfer, this can be achieved either through labeled training examples, as in CTRL \citep{keskar2019ctrl} and PPLM \citep{dathathri20}, or a few-shot approach, as in CoCon \citep{chan20}. These models differ from style transfer models in that they aim to generate plausible \emph{continuations} following a prompt, as opposed to transferring attributes of a fully-formed input while preserving as much content as possible. It is not clear if controlled generation models could be used to perform style transfer, and they have not to our knowledge been evaluated in this context.

\section{Conclusion}
We have presented a unique approach to few-shot text style transfer that is competitive with systems trained with labels (an easier setting), while allowing control of how much of the input is changed. We demonstrate that this approach can produce a single system capable of transferring many different styles while requiring only a handful of exemplars at inference time.

\subsubsection*{Acknowledgments}

We thank Llion Jones, Rami Al-Rfou, and Daniel Gildea for helpful discussion and comments on an earlier draft.

\newpage
\bibliographystyle{acl_natbib}
\bibliography{anthology,acl2021}

\newpage
\phantom{blah}
\newpage
\appendix

\section{Appendix}

\subsection{Beyond Style Transfer}\label{sec:appendix_beyond}

In this section, we provide additional examples illustrating the abilities of our TextSETTR model trained on Common Crawl data, beyond typical style transfer.

Examples of \textbf{shortening} are given in Table~\ref{tab:shortening}, with inputs taken from the first five sentences of the Wikipedia article ``Artificial neural network''. As shortening may require minor rephrases, we set our tuning ranges to add:0--5\%, delete:40--90\%. Since our intention is to leave the style unchanged (apart from length), we extract the target style directly from the input text, with no delta added. The model is largely successful at identifying and removing ``superfluous'' content, and finding ways of rephrasing to shorten while preserving meaning.

\begin{table*}[ht]

\fontsize{8}{9.9}\selectfont
\setlength{\tabcolsep}{0.5em}
\centering
\begin{tabular}{p{.98\textwidth}} \toprule

\hangindent1.47em Artificial neural networks (ANN) \red{\underline{or connectionist systems}} are computing systems that are inspired by\red{\underline{, but not identical to,}} biological neural networks that constitute animal brains. \\

\hangindent1.47em $\Rightarrow$ Artificial neural networks (ANNs) are computing systems that are inspired by \blue{\underline{the}} biological neural networks that constitute animal brains.
\\
\midrule

\hangindent1.47em Such systems \red{\underline{``}}learn\red{\underline{''}} to perform tasks by considering examples, generally without \red{\underline{being programmed with task-specific}} rules.
\\

\hangindent1.47em $\Rightarrow$ Such systems learn to perform tasks by considering examples, generally without \blue{\underline{explicit}} rules.
\\
\midrule

\hangindent1.47em For example, \red{\underline{in}} image recognition\red{\underline{, they}} might learn to identify images that contain cats by analyzing \red{\underline{example}} images that have been manually \red{\underline{labeled}} as ``cat'' or ``no cat'' \red{\underline{and using the results to identify cats in other images}}.
\\

\hangindent1.47em $\Rightarrow$ For example, image recognition \blue{\underline{systems}} might learn to identify images that contain cats by analyzing images that have been manually \blue{\underline{classified}} as ``cat'' or ``no cat''.
\\
\midrule

\hangindent1.47em They do \red{\underline{this without any prior knowledge of}} cats\red{\underline{, for example, that they}} have fur, tails, whiskers and cat-like faces.\\

\hangindent1.47em $\Rightarrow$ They do \blue{\underline{not know that}} cats have fur, tails, whiskers and cat-like faces.
\\
\midrule

\hangindent1.47em Instead, they automatically generate identifying characteristics \red{\underline{from the examples that they process}}.

\hangindent1.47em $\Rightarrow$ Instead, they automatically generate identifying characteristics.
\\
\bottomrule

\end{tabular}
\caption{Examples of shortening (add:0--5\%, delete:40-90\%), using the first five sentences from the Wikipedia article ``Artificial neural network''. For each sentence, the target style is extracted directly from the input text, and no delta is added.}
\label{tab:shortening}
\end{table*}

\begin{table*}[ht]
\fontsize{8}{9.9}\selectfont
\setlength{\tabcolsep}{0.5em}
\centering
\begin{tabular}{p{0.4\textwidth}|p{0.4\textwidth}}

\toprule

\textbf{Add/Delete: 10--30\%} & \textbf{Add/Delete: 30--50\%} \\ \hline

What'll the weather be like? &
What's the weather like? \\

What'll the weather be like tomorrow? &
What will the weather be like tomorrow? \\

What's the weather like tomorrow? &
Will the weather be better tomorrow? \\

What'll the weather be tomorrow? &
What's the weather forecast for tomorrow? \\

What's the weather supposed to be tomorrow? &
How will the weather be tomorrow? \\

\midrule

\textbf{Add/Delete: 50--70\%} & \textbf{Add/Delete: 70--90\%} \\ \hline

Will the weather be perfect tomorrow? &
How do you know what the weather will be like? \\

What's the weather for tomorrow? &
Is it supposed to be cold tomorrow? \\

What's the weather like on the course? &
What will the weather be like in the South? \\

Hopefully the weather will be better tomorrow. &
I'm not a fan of the weather. \\

What's the weather like for the next day? &
What is the temperature and what is the humidity. \\

\bottomrule

\end{tabular}
\caption{Random augmentations of input text ``What'll the weather be tomorrow?'', using random style vector deltas with components sampled from $\mathcal{N}(0, 0.08)$.}
\label{tab:augmentations}
\end{table*}

Examples of \textbf{random augmentations} are given in Table~\ref{tab:augmentations}. In each case, we transfer the input sentence ``What'll the weather be tomorrow?'' to a slightly different style. Specifically, for each transfer, we extract this sentence's style vector and apply a small amount of noise, with each component of the noise vector sampled from a Gaussian $\mathcal{N}(0, 0.08)$. Note that apart from the noise in the style vector, the transfer process is deterministic, as we use greedy decoding.

The cells of Table~\ref{tab:augmentations} apply different tuning ranges, conditioning the model to change a little or a lot. Within each cell, we repeatedly sample the noised style, and present the first five unique outputs. The results indicate that many random changes in style are largely meaning preserving, especially when a small change is requested. With larger add/delete rates, the outputs are still closely related in meaning, despite low lexical overlap.

\subsection{Settings used for Qualitative Analysis}\label{sec:appendix_settings}

For each of the transfer types (e.g.,~formal~$\leftrightarrow$\,informal) in Table~\ref{tab:cherrypicks}, we specify the intended target styles through a tiny set of exemplars. These exemplars are provided in Tables~\ref{tab:emotiveness}--\ref{tab:sentiment}. Additionally, for each transfer type, we select a delta scale $\lambda$ and add/delete rates. These settings are selected through initial experiments, and are held fixed across all examples of transfer shown.

\setlist[enumerate]{leftmargin=1.5em, itemsep=1pt, topsep=0pt, partopsep=0pt}

\begin{table*}[tp!]
\fontsize{8}{9.9}\selectfont
\centering
\begin{tabular}{p{0.47\textwidth}|p{0.47\textwidth}}
\toprule
\textbf{Reserved Exemplars} & \textbf{Emotive Exemplars} \\
\midrule
\begin{enumerate}
    \item That is a very pretty painting.
    \item I'm excited to see the show.
    \item I'm surprised they rescheduled the meeting.
    \item This specimen is an example of the baroque style.
    \item After the performance, we ate a meal.
\end{enumerate} & 
\begin{enumerate}
    \item  OMG, that's such a beautiful painting!
    \item I'm sooo excited to see the show, it's going to be stellar!!
    \item I absolutely can not believe that they rescheduled the meeting!
    \item This wonderful specimen is a truly spectacular example of the baroque style.
    \item After the superb performance, we ate a delicious meal.
\end{enumerate} \\
\bottomrule
\end{tabular}
\caption{Emotiveness transfer exemplars. Transfer settings: $\lambda$=9, add/delete rates: 0--100\%.}
\label{tab:emotiveness}
\end{table*}

\begin{table*}[tp!]
\fontsize{8}{9.9}\selectfont
\centering
\begin{tabular}{p{0.47\textwidth}|p{0.47\textwidth}}
\toprule
\textbf{American Exemplars} & \textbf{British Exemplars} \\
\midrule
\begin{enumerate}
    \item It cost ten bucks. \item My neighbor apologized. \item I'm heading out to the bar with some friends.
\end{enumerate} & 
\begin{enumerate}
    \item It cost ten quid. \item My neighbour apologised. \item I'm heading out to the pub with some mates.
\end{enumerate} \\
\bottomrule
\end{tabular}
\caption{Dialect transfer exemplars. Transfer settings: $\lambda$=8, add/delete rates: 10--30\%.}
\label{tab:dialect}
\end{table*}

\begin{table*}[tp!]
\fontsize{8}{9.9}\selectfont
\centering
\begin{tabular}{p{0.47\textwidth}|p{0.47\textwidth}}
\toprule
\textbf{Polite Exemplars} & \textbf{Rude Exemplars} \\
\midrule
\begin{enumerate}
    \item No thank you, I'd prefer not to. \item This game could have been better designed. \item Do you know why they might have delayed the launch? \item Sorry, I wasn't certain if you were joking.
\end{enumerate} & 
\begin{enumerate}
    \item Hell no, you can't make me do that. \item This game is such a piece of garbage! \item Why in god's name would they delay the damn launch? \item Are you frigging kidding me?
\end{enumerate} \\
\bottomrule
\end{tabular}
\caption{Politeness transfer exemplars. Transfer settings: $\lambda$=5, add/delete rates: 20--50\%.}
\label{tab:politeness}
\end{table*}

\begin{table*}[tp!]
\fontsize{8}{9.9}\selectfont
\centering
\begin{tabular}{p{0.47\textwidth}|p{0.47\textwidth}}
\toprule
\textbf{Formal Exemplars} & \textbf{Informal Exemplars} \\
\midrule
\begin{enumerate}
    \item This was a remarkably thought-provoking read. \item It is certainly amongst my favorites. \item We humbly request your presence at our gala on the 12th.
\end{enumerate} & 
\begin{enumerate}
    \item reading this rly makes u think \item Its def one of my favs \item come swing by our bbq next week if ya can make it 
\end{enumerate} \\
\bottomrule
\end{tabular}
\caption{Formality transfer exemplars. Transfer settings: $\lambda$=4, add/delete rates: 40--80\%.}
\label{tab:formality}
\end{table*}

\begin{table*}[tp!]
\fontsize{8}{9.9}\selectfont
\centering
\begin{tabular}{p{0.47\textwidth}|p{0.47\textwidth}}
\toprule
\textbf{Positive Exemplars} & \textbf{Negative Exemplars} \\
\midrule
\begin{enumerate}
    \item Five stars, I love it.
\end{enumerate} & 
\begin{enumerate}
    \item Zero stars, I hate it.
\end{enumerate} \\
\bottomrule
\end{tabular}
\caption{Sentiment transfer exemplars. Transfer settings: $\lambda$=3, add/delete rates: 0--100\%.}
\label{tab:sentiment}
\end{table*}

\subsection{Human Reference BLEU}\label{sec:reference_bleu}

\begin{figure*}
  \centering
  \begin{minipage}[c]{0.44\textwidth}
    \centering
    \setlength{\tabcolsep}{0.25em}
    \resizebox{0.8\columnwidth}{!}{
    \footnotesize
    \begin{tabular}{ccc}
      \toprule
      \textbf{Model} & \textbf{BLEU} & \textbf{Self-BLEU} \\
      \midrule
      CrossAligned & 2.0 & 2.9\\
      Delete\&Retrieve & 29.7 & 56.9\\
      B-GST & 29.0 & 54.2\\ \hline
      CP-G & 17.0 & 35.5\\
      CP-B & 19.4 & 39.8\\
      TextSETTR (0--20\%) & 39.0 & 73.3\\
      TextSETTR (10--30\%) &  30.7 & 55.8\\
      TextSETTR (20--40\%) & 20.0 & 34.7\\
      TextSETTR (30--50\%) & 10.6 & 18.4\\
      TextSETTR (40--60\%) & 5.5 & 9.1\\
      TextSETTR (50--70\%) & 2.2 & 3.6\\
      \bottomrule
    \end{tabular}}

  \end{minipage}
  \begin{minipage}[c]{0.5\textwidth}

    \centering
    \includegraphics[trim=0 2 0 16,clip,scale=0.6]{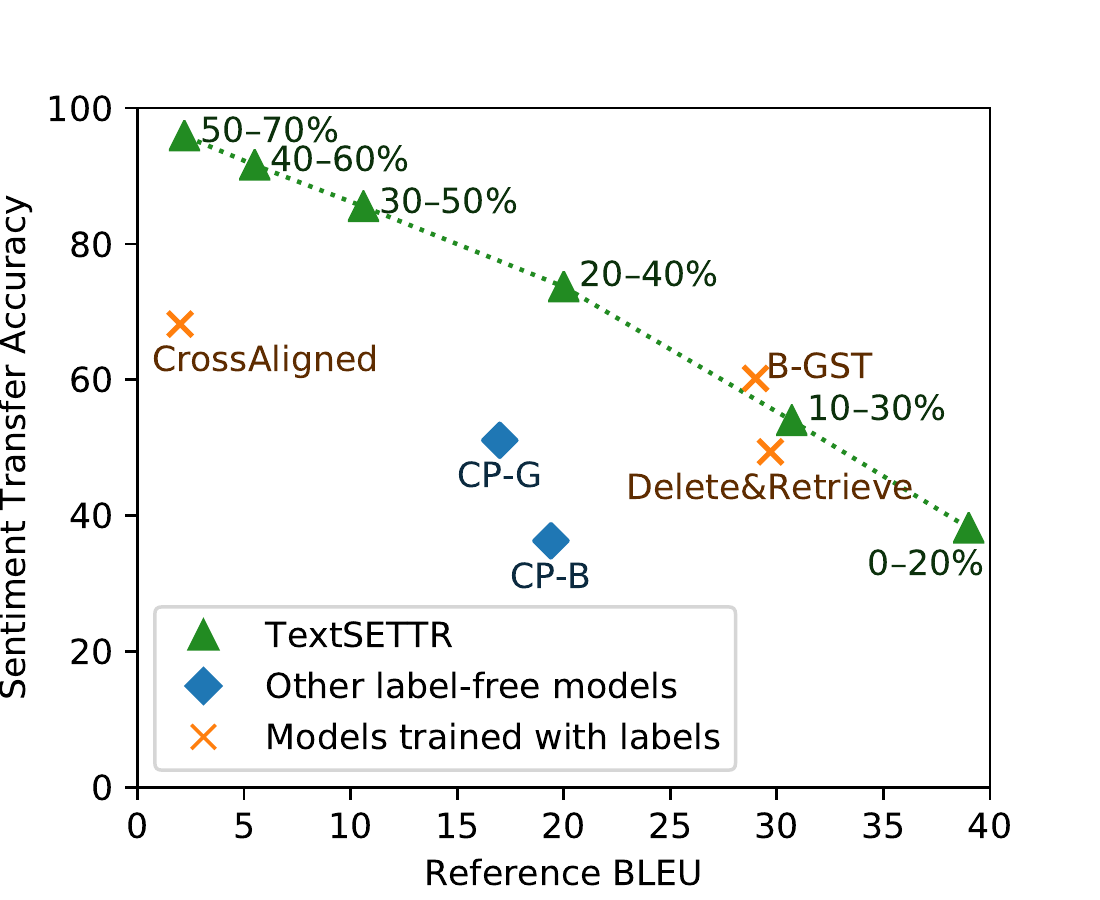}

  \end{minipage}
  \caption{BLEU scores between model outputs and human references provided by \citet{li18}, along with self-BLEU for comparison. The first group of models in the table had access to labels at training time, while the second group did not. TextSETTR (X--Y\%) refers to our model with add/delete rate ranges set to X--Y\%.}\label{fig:reference_bleu}
\end{figure*}

\citet{li18} provide human reference transfers for their Amazon test data, and report BLEU scores of model outputs against these targets. In principle, we believe this metric is less informative than self-BLEU, as style transfer is a relatively open-ended task, and successful transfers may differ significantly from the single human reference. However, for completeness, we report ``reference BLEU'' of our models and those of prior work in Figure~\ref{fig:reference_bleu}. We observe BLEU and self-BLEU are highly correlated, and the ``Accuracy vs.~BLEU'' plot conveys the same relationships we saw in Figure~\ref{fig:results}. As before, all BLEU scores are calculated using SacreBLEU \citep{post18}.

\subsection{Amazon Reviews Preprocessing}\label{sec:appendix_preprocessing}

We use the code in Figure~\ref{fig:preprocessing} to process raw Amazon reviews from the \citet{ni-etal-2019-justifying} dataset and extract pairs of adjacent lines, preprocessed to have a similar format to \citet{li18} dataset. We split reviews on newlines, and clip lines to 100 characters, always ending with a period. This gives results similar to \citet{li18}, where one line may contain multiple sentences, and may consists of a ``half-sentence'' ending with ``e.g.'' or a similar non-sentence-final period. Additionally, we apply various tokenization and normalization operations to roughly match the observed \citet{li18} text.

\begin{figure*}
\footnotesize
\begin{verbatim}
import re
from html.parser import HTMLParser

html_parser = HTMLParser()

def preprocess(line):
  """Simulate Li et al. preprocessing of one review line."""
  # Lowercase.
  line = line.lower()
  # Replace apostrophes, parens and quotes with spaces.
  line = re.sub("['()\"]", " ", line)
  # Replace dollar values ==> $
  line = re.sub("\$[\d.]*", "$", line)
  # Replace percent values ==> %
  line = re.sub("[\d.]*%", "%", line)
  # Replace single digits ==> num_num
  line = re.sub(" \d[ ,]", " num_num ", line)
  # Replace multi-digits and codes ==> num_extend 
  line = re.sub(" \d[^ ]*", " num_extend", line)
  # Remove remaining numbers, including decimals.
  line = re.sub("\d[\d.]*", "", line)
  # Add spaces around certain punctuation marks.
  line = re.sub("([.,?!:])", r" \1 ", line)
  # Remove double spaces after periods before words.
  return re.sub(r"\.  ([a-z])", r". \1", line)

def acceptable_line(line):
  """Check if text looks like an acceptable line from Li et al."""
  if not line or len(line) < 30 or len(line) >= 100:
    return False
  # Avoid lines with any char absent from Li et al. train.
  if re.search('[^ !$%+,.:;>?@\^_`a-z{|}]', line):
    return False
  return True

def clip_to_last_period(line):
  return line[:len(line) - line[::-1].index('.')]

def adjacent_lines(review):
  """Extract a list of adjacent line pairs from review text."""
  review = html_parser.unescape(review)
  review = review.replace('\\"', '"')
  # Simulate Li et al. splitting and filtering.
  if '\n' not in review:
    return  
  lines = review.split('\n')
  lines = [preprocess(clip_to_last_period(l[:100]))
           for l in lines if l and "." in l[:100]]
  lines = [preprocess(l) for l in lines]
  lines = [l for l in lines if acceptable_line(l)]
  if len(lines) < 2:
    return
  return list(zip(lines[:-1], lines[1:]))
\end{verbatim}
\caption{Python code to extract adjacent lines of text from raw Amazon reviews, producing outputs in a similar style to \citet{li18}.}
\label{fig:preprocessing}
\end{figure*}

\subsection{Human Evaluation Setup}\label{sec:appendix_human}

For the human evaluations of our models, we employed 3 in-house annotators.
The annotators were paid hourly wages that are competitive for their locale and have standard rights as contractors.
They spoke native English.

For the evaluation task, the annotators were shown both the original and transformed pieces of text.
They were then asked to evaluate for three metrics: fluency, meaning preservation, and sentiment change.

For \emph{fluency}, they were asked, ``For the new text, how do you rate the fluency, i.e., the quality and readability of the text, with 1 being not fluent at all and 5 being very fluent.''
For \emph{meaning preservation}, they were asked, ``Comparing the new text against the original text, and ignoring the change of style, how well does the new text preserve as much of the original meaning, with 1 being all meaning is lost and 5 being preserving as much as possible given the sentiment change?''
And for \emph{sentiment change}, they were asked, ``Comparing the new text against the original text, how well did the sentiment of the new text become more positive, with 1 being not more positive and 5 being a lot more positive?''

\end{document}